\pdfoutput=1
\documentclass[11pt]{article}

\usepackage[]{emnlp2021}

\usepackage{times}
\usepackage{latexsym}

\usepackage[T1]{fontenc}
\usepackage[utf8]{inputenc}

\usepackage{microtype}
\usepackage{listings}
\usepackage{tikz}

\def\checkmark{\tikz\fill[scale=0.4](0,.35) -- (.25,0) -- (1,.7) -- (.25,.15) -- cycle;}

\title{BNLP: Natural language processing toolkit for Bengali}

\author{Sagor Sarker \\
  Begum Rokeya University, Rangpur, Bangladesh \\
  \texttt{brursagor@gmail.com} \\}

\begin{document}
\maketitle
\begin{abstract}
BNLP is an open-source language processing toolkit for Bengali consisting of tokenization, word embedding, part of speech(POS) tagging, name entity recognition(NER) facilities. BNLP provides pre-trained model with high accuracy to do model-based tokenization, embedding, POS, NER tasks for Bengali. BNLP pre-trained model achieves significant results in Bengali text tokenization, word embeddings, POS, and NER task. BNLP is being used widely by the Bengali research communities with 25K downloads, 138 stars, and 31 forks. BNLP is available at \url{https://github.com/sagorbrur/bnlp}.
\end{abstract}

\section{Introduction}
Natural language processing is one of the most important fields in computation linguistics. Tokenization, embedding, POS, NER, text classification, language modeling are some of the sub-tasks of NLP. Any computational linguistics researcher or developer needs hands-on tools to do these subtasks efficiently. Due to the recent advancement of NLP, there are so many tools and methods to do word tokenization, word embedding, POS, NER in the English language. NLTK \cite{loper}, coreNLP \cite{manning_2014}, spaCy \cite{spacy2}, AllenNLP \cite{DBLP:journals/corr/abs-1803-07640}, Flair \cite{akbik_2019}, stanza \cite{qi2020stanza} are few of the tools. These tools provide a variety of methods to do tokenization, embedding, POS, NER, language modeling for the English language. Support for other low resource languages like Bengali is limited or no support at all. A recent tool like iNLTK \cite{arora_2020} is an initial approach for different Indic languages including Bengali. But as it groups with other indic languages special monolingual support like easy pre-processing, tokenization, embedding, POS, NER for Bengali is missing. Besides, iNLTK is mostly based on deep learning(DL) language model based pipeline, which needs DL based infrastructure to do NLP tasks. And that makes iNLTK verbose and language model centric tool for Bengali language. On the other side, BNLP is totally machine learning(ML) based toolkit that can do an instant process for Bengali NLP tasks. Table \ref{tab:compare} provides detailed feature comparison between BNLP and other tools.

% \begin{figure}
%     \centering
%     \includegraphics[width=0.50\textwidth]{assets/bnlp.png}
%     \caption{BNLP{}'s Feature}
%     \label{fig:architecture}
% \end{figure}

\begin{table*}[]
    \centering
        \begin{tabular}{ccccccccc}
        \hline
        Tool & \begin{tabular}[c]{@{}c@{}}Support\\  Bengali\end{tabular} & ML Based & \begin{tabular}[c]{@{}c@{}}Pre-trained\\  Model\end{tabular} & Tokenizer & Embedding & POS & NER & LM \\
        \hline
        NLTK &  & \checkmark &  &  &  &  & & \\
        spaCy & \checkmark &  &  &  &  &  & &\\
        Flair &  &  &  &  &  &  & &\\
        stanza &  &  &  &  &  &  & \\
        inltk & \checkmark &  & \checkmark & \checkmark &  &  & & \checkmark\\\hline
        BNLP & \checkmark & \checkmark & \checkmark & \checkmark & \checkmark & \checkmark & \checkmark\\\hline
        \end{tabular}
    \caption{BNLP feature comparison with other popular tools}
    \label{tab:compare}
\end{table*}

% \checkmark

BNLP is an open-source language processing toolkit for Bengali is build to address this problem and breaks the barrier to do different Bengali NLP tasks by:
\begin{itemize}
    \item Providing different tokenization methods to tokenize Bengali text efficiently
    \item Providing different embedding method to embed Bengali word using the pre-trained model and also provides an option to train an embedding model from scratch
    \item Providing hands-on start option for POS or NER of Bengali sentences and also provides an option for training CRF based POS tagger or NER model from scratch.
\end{itemize}

BNLP offers several widely used text preprocessing techniques like removing stopwords, removing punctuations, removing foreign words. BNLP Github repositories\footnote{\url{https://github.com/sagorbrur/bnlp}} for source code of the package, pre-trained model and documentation\footnote{\url{https://bnlp.readthedocs.io/}}. BNLP libraries have a permissive MIT license. BNLP is easy to install via pip or by cloning repository, easy to plugin with any python projects. 

\section{Related Works}
There is a significant number of open-source NLP tools for the English language. Tools like NLTK \cite{loper}, coreNLP \cite{manning_2014}, spaCy \cite{spacy2}, AllenNLP \cite{DBLP:journals/corr/abs-1803-07640}, Flair\cite{akbik_2019}, stanza \cite{qi2020stanza} are few of the tools. These tools mostly build for the English language and have limited or no support for low resource languages. Especially in a low resource language like Bengali, there is a huge scarcity of tools to process. iNLTK \cite{arora_2020} is an initial approach to help process Bengali with tokenization, language model support. But as it's a group with different Indic languages, a special monolingual concern for Bengali is missing. Keeping that concern in mind we build BNLP to support especially Bengali and provides tokenization, embedding, POS, NER supports. Besides, iNLTK is mostly based on DL language model based pipeline, which needs DL based infrastructure to do NLP tasks. And that makes iNLTK verbose and language model centric tool for Bengali language. On the other side, BNLP is a totally ML based toolkit that can do an instant process for Bengali NLP tasks.
% inltk only support lm based support which mostly depend on deep learning framework like pytorch which is heavy for instant inference
% inltk didn't provide any basic tokenizer support beside sentencepiece
% inltk didn't provide pos, ner support
% need to provide strong support that confirm that bnlp is better than inltk, inference time, tokenizer, pos, ner will be better choice
% add other tool like bnlm, or Bengali-bert to differentiate your tool from other tool

\section{BNLP API}
Our design principle was to make the tool easily usable with a few lines of code. The researcher or developer can integrate this tool with installing a simple python package.
In this section, we are describing how to do different NLP tasks for Bengali text using BNLP toolkit.

\begin{figure}
    \centering
    \includegraphics[width=0.48\textwidth]{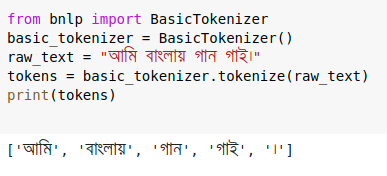}
    \caption{An example of doing basic tokenization using BNLP}
    \label{fig:basict}
\end{figure}

\begin{figure}
    \centering
    \includegraphics[width=0.48\textwidth]{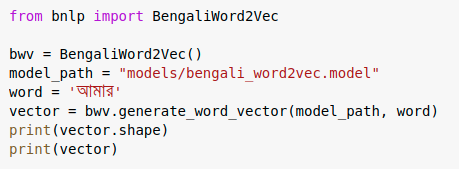}
    \caption{An example of generating word vector using trained model and BNLP}
    \label{fig:w2v}
\end{figure}

% \begin{lstlisting}[language=python]
% from bnlp import BasicTokenizer
% basict = BasicTokenizer()
% basict.tokenize(text)
% \end{lstlisting}

\begin{table*}[]
\centering
\begin{tabular}{cccc}
\hline
\textbf{Corpus} & \textbf{Articles} & \textbf{Sentences} & \textbf{Tokens} \\
\hline
Wikipedia & 99139 & 1818523 & 32908419\\
News Articles & 127867 & 4017940 & 60526710\\\hline
Total & 227006 & 5836463 & 93435129\\\hline
\end{tabular}
\caption{Statistics of Datasets used for training sentencepiece, word2vec, fasttext Models}
\label{tab:rawdata}
\end{table*}

\begin{table}
\centering
\begin{tabular}{cccc}
\hline \textbf{} & \textbf{Sentences} & \textbf{Train} & \textbf{Test} \\ \hline
POS & 2997 & 2247 & 750 \\
NER & 67719 & 64155 & 3564 \\
\hline
\end{tabular}
\caption{\label{seqdata} Statistics of POS and NER datasets }
\end{table}

\begin{table}
\centering
\begin{tabular}{cccc}
\hline \textbf{Task} & \textbf{Precision} & \textbf{Recall} & \textbf{F1} \\ \hline
POS & 81.74 & 79.78 & 80.75 \\
NER & 74.15 & 60.91 & 66.88 \\
\hline
\end{tabular}
\caption{\label{posner_eval} Evaluation results}
\end{table}

\subsection{Tokenizers}
BNLP provides three different tokenization options to tokenize Bengali text. Under rule-based tokenizer BNLP provides \textbf{Basic Tokenizer} a punctuation splitting tokenizer and \textbf{NLTK}\footnote{\url{https://github.com/nltk/nltk}} tokenizer. As NLTK tokenizer is for the English language, we modified nltk tokenize output to use it for Bengali keeping in mind the difference between punctuation of English and Bengali. Under model-based tokenization BNLP provides \textbf{sentencepice}\footnote{\url{https://github.com/google/sentencepiece}} tokenizer for Bengali text called Bengali Sentencepiece. Bengali sentencepiece API provides two options, the pretrained sentencepiece model and the training sentencepiece model. Anyone can tokenize Bengali text using a pretrained sentencepiece model or can train their own Bengali sentencepiece model by calling train API. Figure \ref{fig:basict} shows an example of BNLP basic tokenizer.

% figure here
\subsection{Embedding}
BNLP provides two different embedding option to embed Bengali words, one is word2vec \cite{DBLP:journals/corr/abs-1301-3781} and another is fasttext \cite{DBLP:journals/corr/BojanowskiGJM16}. Both Bengali word2vec and fasttext has two option, one is embed Bengali word using pre-trained model and another is training Bengali word2vec or fasttext model from scratch. For both embedding model, we used gensim\footnote{\url{https://github.com/RaRe-Technologies/gensim}} embedding API and trained with Bengali corpora. Figure \ref{fig:w2v} shows an example of generating word vector using pre-trained model and BNLP.

\subsection{POS Tagging}
BNLP provides a hands-on starting option for POS to Bengali by giving a method to tag part of speech from a given sentence using pre-trained CRF based \cite{mccallum_li_2003} model. BNLP also provides an option to train a CRF-based POS model with custom POS datasets.
% \begin{figure}
%     \centering
%     \includegraphics[width=0.48\textwidth]{assets/pos.png}
%     \caption{BNLP POS API}
%     \label{fig:pos_api}
% \end{figure}

\subsection{NER Tagging}
\begin{figure}
    \centering
    \includegraphics[width=0.48\textwidth]{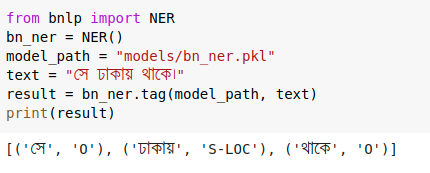}
    \caption{An example of doing NER using BNLP}
    \label{fig:ner_api}
\end{figure}

BNLP provides a hands-on starting option for NER to Bengali by giving a method to tag name entity from a given sentence using a pre-trained CRF-based NER model. BNLP also provides an option to train a CRF-based NER model with custom NER datasets. Figure \ref{fig:ner_api} shows an example of name entity tagging using BNLP.

Apart from this BNLP provides some extra utilities methods like getting Bengali \textbf{stopwords}, \textbf{letters}, \textbf{punctuation} from \textbf{Corpus} class. 

\section{Pre-trained Models}
BNLP provides different pre-trained Bengali model including (i) sentencepiece (ii) word2vec (iii) fasttext (iv) CRF-based POS tagging (v) CRF-based NER tagging model.

\textbf{Sentencepiece}: For training different language models we need subword level better vocabulary. We build subword-based vocabulary by training sentencepice model on Bengali Wikipedia and news articles datasets. We trained sentencepiece unigram language model \cite{DBLP:journals/corr/abs-1804-10959} with vocab size of 50000.

\textbf{Word2Vec}: We trained Bengali word2vec model on  Bengali Wikipedia and news articles datasets using gensim word2vec pipeline. We trained our word2vec model with embedding dimension 300, window size 5, the minimum number of word occurrences 1, and total workers number 8. We train it for a total of 50000 iterations.

\textbf{Fasttext}: We trained Bengali Fasttext model on Bengali Wikipedia and news articles datasets. For training fasttext we set embedding dimension 300, windows size 5, number of minimum word occurrences 1, model type skip-gram, learning rate 0.05. We trained a total of 50 epochs and our loss is 0.318668.

\textbf{CRF-Based POS Tagging Model}: We trained our CRF-Based POS tagging model on nltr \footnote{\url{https://github.com/abhishekgupta92/bangla_pos_tagger}} datasets. We split data into 75\% train and 25\% test. Our evaluation result for the POS tagging model is 80.75 F1 score. 

\textbf{CRF-Based NER Model}: We trained our CRF-Based NER model on NER-Bengali-Datasets \cite{karim_et_al2019}. We split data into 75\% train and 25\% test. Our evaluation result for the NER model is 68.88 F1 score. 

Table \ref{seqdata} provides detailed evaluation results of POS tagging and NER model.

\subsection{Datasets}
For training sentencepiece, word2vec, fasttext we used Bengali raw text data from two sources. One is wikipedia\footnote{\url{https://dumps.wikimedia.org/bnwiki/latest/}} dump dataset and another is crawl news articles from different news portal sites. As shown in Table \ref{tab:rawdata} our raw data contains a total of 99139 Wikipedia Bengali articles and 127867 news articles. Wikipedia corpus contains a total of 1818523 sentences with 32908419 tokens. News articles corpus contains a total of 4017940 sentences with 60526710 tokens.

For POS tagging we used \textbf{nltr} datasets which contains total of 2997 sentences. We split those datasets into 2247 train and 750 test sets and train our POS tagging model. For NER we used NER-Bengali-Datasets \cite{karim_et_al2019} which contains a total of 67719 data with 64155 train and 3564 test. Table \ref{seqdata} provides details statistics of POS and NER datasets.

% \subsection{Training and Evaluation}
% We train sentencepiece model with our raw text data with vocab size 50000. As sentencepiece provide us end-to-end system for training and tokenizing we did not do any preprocessing task in our raw text datasets.

% We train our word2vec model with embedding dimension 300, window size 5, the minimum number of word occurrences 1, and total workers number 8. We train it for a total of 50000 iterations.

% For training fasttext we set embedding dimension 300, windows size 5, number of minimum word occurrences 1, model type skip-gram, learning rate 0.05. We trained a total of 50 epochs and our loss is 0.318668.

% Our CRF-based POS tagging model and NER tagging model training approach is similar. We split data into 75\% train and 25\% test. Our evaluation result for the POS tagging model is 80.75 F1 score and the NER model is 66.88 F1 score. Table \ref{posner_eval} describe details about evaluation results.

\section{Conclusion and Future Work}
BNLP language processing toolkit provides tokenization, embedding, POS, NER facilities for Bengali. BNLP pre-trained model achieves significant results in Bengali text tokenizing, word embedding, POS, and NER task. BNLP is being used widely by Bengali research communities and appreciated by the communities.

We are working on extending the support tools like stemming, lemmatizing, corpus support for BNLP in the future. We are working on adding language model-based support in BNLP so that researchers can use it for different downstream tasks efficiently. While these tasks under development, we are hoping that BNLP will accelerate Bengali NLP research and development.

\bibliography{anthology,custom}
\bibliographystyle{acl_natbib}

% \appendix

% \section{Example Appendix}
% \label{sec:appendix}

% This is an appendix.

\end{document}